# Probabilistic Latent Semantic Analysis (PLSA) untuk Klasifikasi Dokumen Teks Berbahasa Indonesia

**DERWIN SUHARTONO**

Technical Report
Fakultas Ilmu Komputer
Program Studi Doktor Ilmu Komputer
Universitas Indonesia
Desember 2014

**Abstrak**

Salah satu pekerjaan yang ada di dalam mengelola dokumen adalah bagaimana menemukan intisari dari dokumen. *Topic modeling* merupakan teknik yang dikembangkan untuk menghasilkan representasi dokumen berupa kata-kata kunci dari dokumen. Kata-kata kunci tersebut yang akan digunakan dalam proses pengindeksan serta pencarian dokumen untuk ditemukan kembali sesuai kebutuhan pengguna. Pada penelitian ini, akan dibahas secara spesifik mengenai Probabilistic Latent Semantic Analysis (PLSA). Pembahasan akan meliputi mekanisme bagaimana PLSA yang juga melibatkan algoritma Expectation Maximization (EM) sebagai pelatihan diterapkan pada sekumpulan data korpus, serta bagaimana melakukan uji coba dan memperoleh akurasi hasil penggunaan PLSA.

## 1. Pendahuluan

Teks merupakan salah satu media yang mampu mengkomunikasikan berbagai macam hal. Apabila dibandingkan dengan gambar, video, audio, animasi, dan lain sebagainya, maka teks merupakan bagian utama yang ada di dalam dokumen. Surat, majalah, dan koran merupakan contoh media yang menggunakan teks dan sudah banyak dikenal dengan baik oleh berbagai kalangan. Jika dilakukan observasi dan diamati lebih lanjut, dokumen teks yang sebelumnya dipublikasikan pada berbagai media cetak mulai beralih menjadi dokumen teks yang dipublikasikan dalam bentuk elektronik. Tentunya dengan pergantian bentuk dokumen tersebut, jumlah dokumen elektronik mengalami peningkatan jumlah yang cukup pesat. Kondisi ini mendorong pada munculnya keperluan akan pengelolaan dokumen elektronik yang baik.

Salah satu pekerjaan yang ada di dalam mengelola dokumen adalah bagaimana menemukan intisari dari dokumen. Hal ini terkait erat dengan melakukan perangkuman atau membuat representasi yang ringkas dari dokumen. Panjang dokumen sangat bervariasi, dari bentuk dokumen yang singkat/pendek seperti memo atau dokumen yang panjang seperti berita elektronik, buku, skripsi, tulisan ilmiah dan sejenisnya. Pencarian dokumen umumnya dilakukan dengan menggunakan kueri berupa kata, frase, kalimat atau judul dokumen yang dibutuhkan. Apabila pencarian dilakukan dengan pencocokan kata per kata dari kueri ke dokumen, tentu akan memakan waktu komputasi yang lama. Hal ini mendorong pada keperluan dalam pembentukan representasi dari dokumen yang menghilangkan kata-kata yang tidak mewakili arti dari dokumen serta mempertahankan kata-kata yang memiliki nilai tinggi sehingga dianggap bisa mewakili arti dari dokumen aslinya.

*Topic modeling* merupakan teknik yang dikembangkan untuk menghasilkan representasi dokumen berupa kata-kata kunci dari dokumen. Kata-kata kunci tersebut yang akan digunakan dalam proses pengindeksan serta pencarian dokumen untuk ditemukan kembali sesuai kebutuhan pengguna. Latent Semantic Analysis (LSA) muncul sebagai teknik pertama yang bisa menghasilkan representasi dokumen berupa kumpulan kata-kata. LSA merupakan metode yang paling banyak dikenal dengan melekatkan ciri Bag-of-Words (Landauer, Foltz, dan Laham, 1998) sebagai representasi dokumen. Probabilistic Latent Semantic Analysis (PLSA) yang dikembangkan oleh Hoffman (1999) merupakan LSA yang menggunakan nilai probabilistik sebagai penentu bobot topik dari setiap dokumen yang ada.Sebagai varian baru dari LSA, teknik GLSA diajukan oleh Islam dan Hoque (2012) yang mengubah keberadaan term di dalam term-document matrix menjadi n-gram. Sedangkan, metode bernama Multidimensional Latent Semantic Analysis (MDLSA) yang diusulkan oleh Zhang, Ho, Wu, dan Ye (2013) merupakan metode yang meninjau kepada hubungan term dan distribusi spasial. Teknik yang melibatkan aspek sintaksis secara langsung adalah SELSA (Syntactically Enhanced Latent Semantic Analysis) yang diajukan oleh Kanejiya, Kumar dan Prasad (2003).

*Topic modeling* masih berupa model sehingga masih luas area cakupan penerapannya dalam berbagai jenis aplikasi. Pada laporan ini, akan dibahas secara spesifik mengenai salah satu teknik *topic modeling* yaitu Probabilistic Latent Semantic Analysis (PLSA). Pembahasan akan meliputi mekanisme bagaimana PLSA yang juga melibatkan algoritma Expectation Maximization (EM) sebagai pelatihan diterapkan pada sekumpulan data korpus, serta bagaimana melakukan uji coba dan memperoleh akurasi hasil penggunaan PLSA. Data korpus yang digunakan adalah data teks bahasa Indonesia yang diperoleh dari beberapa sumber yakni Kompas, Tempo, dan Republika. Hasil akurasi diukur dengan menggunakan tool untuk klasifikasi dengan membandingkan matriks hasil training data dengan matriks dari sekumpulan data yang hendak diujicobakan. Kemudian, pada bagian akhir dari laporan ini dikemukakan kesimpulan yang diperoleh serta saran untuk penelitian serta eksperimen selanjutnya.

## 2. Landasan Teori

### 2.1. Probabilistic Latent Semantic Analysis

Sebelum kemunculan Probabilistic Latent Semantic Analysis (PLSA) pada tahun 1999, 1 (satu) tahun sebelumnya, Landauer, Foltz dan Laham (1998) menciptakan satu teknik yang dinamakan Latent Semantic Analysis (LSA). LSA adalah sebuah teknik statistik terotomasi untuk membandingkan kesamaan semantik dari beberapa kata atau beberapa dokumen. LSA bukanlah sebuah program yang tradisional dari *natural language processing* ataupun *artificial intelligence*. Teknik ini digunakan dalam menganalisis dokumen untuk menemukan arti atau konsep dari dokumen tersebut. LSA lahir karena didasarkan pada pemikiran bahwa *syntax* dan *style* saja tidak mencukupi untuk menilai sebuah esai. Yang menjadi kesulitan mendasar adalah ketika kita hendak membandingkan kata-kata untuk menemukan dokumen yang relevan. Sebenarnya yang dibandingkan adalah arti atau konsep di balik kata-kata tersebut. Metode LSA melakukan *mapping* dari kata ataupun dokumen menjadi sebuah *concept space* dan perbandingan dilakukan pada *space* ini. *Concept space* tersebut atau yang lebih sering disebut sebagai *latent semantic space* merupakan hasil *mapping* dari matriks dimensi tinggi menjadi dimensi yang lebih kecil. Meskipun dalam dimensi yang lebih kecil, matriks tersebut merupakan matriks yang merepresentasikan isi dari keseluruhan dokumen. Ciri khas dari LSA adalah teknik yang dinamakan Singular Value Decomposition (SVD). SVD digunakan untuk melakukan dekomposisi matriks setelah diberikan pembobotan untuk kemudian diukur kesamaannya dengan data yang akan diujicobakan.

Inti pekerjaan dari *topic modeling* adalah menghasilkan representasi dokumen yang baik sehingga dapat digunakan untuk berbagai macam task. Hofmann(1999) mengemukakan bahwa titik permulaan dari ide PLSA adalah pada sebuah model statistik yang disebut sebagai *aspect model*. Sebuah dokumen teks terdiri dari kumpulan kata-kata. Apabila ditarik satu hal diantara dokumen dan kata tersebut, bisa diperoleh kata kunci (*keyword*) yang menjadi jembatan diantara dokumen dan kata. Kata kunci ini yang disebut sebagai *aspect model*. *Aspect model* didefinisikan sebagai sebuah variabel yang tidak terlihat (*latent variable*) dari sebuah dokumen.

Semua variabel yang digunakan untuk memodelkan aspect model tersebut melibatkan asumsi *conditional independence*. Dokumen dan kata dalam kondisi *conditional independence* dihubungkan dengan topik seperti tergambarkan pada gambar 1. *Joint probability* antara dokumen (D) dengan kata (W) itu tergambarkan pada persamaan berikut:

$$P(d,w) = P(d)P(w|d), \quad P(w|d) = \sum_{z \in Z} P(w|z)P(z|d)$$

Model tersebutbisa digambarkan juga dengan persamaan:

$$P(d,w) = \sum_{z \in \mathcal{Z}} P(z)P(d|z)P(w|z)$$

Semua variabel yang ada di atas merupakan matriks probabilitas. Matriks yang digunakan pada PLSA tidak sama seperti *term-documentmatrix* yang ada pada LSA. Matriks yang digunakan pada PLSA sudah merupakan *likelihood* dari faktor topik terkait dengan dokumen (*document given topic*) dan juga terkait dengan kata (*word given topic*), sedangkan kebalikannya pada LSA, fungsi matriksnya hanya berfungsi sebagai *likelihood term-frequency*.

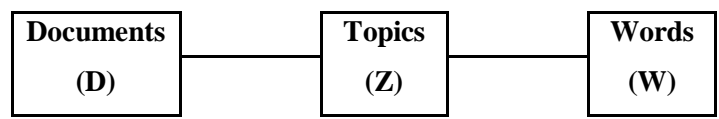

**Gambar 1. Hubungan antar Dokumen, Topik, dan Kata**

Kehadiran topik (Z) di antara kata (W) dan dokumen (D) sebenarnya menyatakan mengapa ketika berada pada dokumen (D), maka kata-kata yang muncul adalah *word* (W). Hal ini menunjukkan adanya sebuah ketergantungan (*dependence*), namun karena ketergantungannya tidak secara langsung, sehingga hal ini dinamakan sebagai sebuah *conditional independence*.

Pembentukan model PLSA untuk menghasilkan kumpulan topik (Z) beserta dengan nilai probabilitasnya diawali dengan pembuatan nilai probabilitas secara acak (random) atau bisa juga dengan dilakukan inisialisasi matriks secara greedy. Pembuatan nilai secara acak ini disebut sebagai proses stokastik karena proses ini mengandung ketidakpastian. Apabila dilakukan inisialisasi secara greedy, hal ini dimaksudkan untuk membentuk *gradient descent* untuk mencapai probabilitas maksimum (*maximum probability*). Setelah diperoleh matriks awal probabilitas, matriks tersebut akan diproses ke dalam training dengan jumlah iterasi tertentu untuk memperoleh probabilitas yang terbaik dengan menggunakan algoritma Expectation Maximization (EM). Algoritma ini terdiri dari 2 tahap yaitu:

1) Expectation (E), dimana probabilitas posterior dihitung untuk latent variable.

    Nilai probabilitas yang diperoleh dari langkah Expectation adalah:

$$P(z|d,w) = \frac{P(z)P(d|z)P(w|z)}{\sum_{z' \in \mathcal{Z}} P(z')P(d|z')P(w|z')}$$

2) Maximization (M), dimana parameter-parameter yang ada akan diperbarui nilainya.

    Nilai yang diperoleh dari langkah Maximization adalah:

$$P(w|z) \propto \sum_{d \in \mathcal{D}} n(d,w) P(z|d,w)$$

$$P(d|z) \propto \sum_{w \in \mathcal{W}} n(d,w) P(z|d,w)$$

$$P(z) \propto \sum_{d \in \mathcal{D}} \sum_{w \in \mathcal{W}} n(d,w) P(z|d,w)$$

Algoritma EMakan mengusahakan untuk memperoleh nilai error yang semakin kecil. Apabila nilai error masih tinggi, nilai bobot akan dioptimasikan supaya nilai error yang ada semakin kecil. Sehingga akan sangat memungkinkan bahwa apabila dilakukan semakin banyak training, maka topik serta probabilitas yang dibentuk akan semakin baik.

Matriks yang dibentuk hasil dari PLSA terdiri dari P(d|z), P(w|z) dan P(z). Matriks P(d|z) adalah matriks probabilitas yang menggambarkan sebaran nilai topik pada sebuah dokumen, sedangkan matriks P(w|z) menggambarkan nilai probabilitas topik pada setiap kumpulan kata-kata, lalu P(z) merupakan nilai probabilitas dari topik itu sendiri.

$$P(d|z) = \begin{pmatrix} & & & & & & \\ & & & & & & \\ & & & & & & \\ & & & & & & \\ & & & & & & \\ z1 & z2 & z3 & z4 & ... & zk \end{pmatrix} \begin{matrix} d1 \\ d2 \\ d3 \\ d4 \\ ... \\ dn \end{matrix}$$

**Gambar 2. Matriks P(d|z)**

Dari skema matriks di atas, bisa dilihat hubungan antara baris dan kolom yaitu bahwa pada setiap dokumen akan memiliki probabilitas topik masing-masing. Sebagai contohnya, $d_1$ (dokumen pertama) memiliki probabilitas dari $z_1$ (topik pertama), $z_2$, hingga $z_n$, begitu juga dengan $d_2$, $d_3$, $d_4$, dan seterusnya.

Data yang digunakan untuk diinputkan ke PLSA adalah dokumen yang menggunakan representasi berupa *term-document matrix*. Misalnya, dimensi awal dari *term-document matrix* adalah n buah dokumen, dan m buah kata. Apabila kita memiliki 100 dokumen dan 1000 kata, maka komponen dari term-document matrix akan berjumlah 100.000. Dari ukuran 100x1000, bisa dibuat distribusi probabilitas P(d|w). Sebagai contoh, dari distribusi data tersebut, apabila hendak diketahui berapa jumlah kata "partai" pada dokumen pertama, maka dapat diperoleh dari sel perpotongan antara dokumen pertama (pada kolom pertama matriks) dengan kata "partai" (pada baris tertentu pada matriks). Hal ini membuat pengolahan data terlihat praktis dan hasilnya pun pasti baik dan relevan, namun permasalahan yang terjadi pada dimensi 100 x 1000 tersebut adalah dimensi matriks yang terlalu besar. Dimensi yang terlalu besar akan memperlambat proses komputasi dan banyak proses yang tidak perlu dilakukan akan tetap dilakukan. Matriks yang terlalu *sparse* akan mengakibatkan pemborosan memori di dalam prosesnya. PLSA menjadi salah satu solusi reduksi dimensi. Hal ini terjadi karena PLSA memiliki jembatan berupa topik Z yang membagi probabilitas menjadi 3 matriks tetapi dengan dimensi yang lebih kecil. Pada proses reduksi dimensi ini, Singular Value Decomposition (SVD) akan berperan secara aktif.

Berikut ini diberikan contoh untuk memudahkan penggambaran reduksi dimensi pada matriks probabilitas PLSA. Misalnya terdapat 100 dokumen dan 1000 kata di dalam dokumen tersebut, kemudian didefinisikan ada 10 topik (Z=10) yang dihasilkan, maka data yang ada menjadi 100 dokumen, 1000 kata, dan 10 topik. Pada term-document matrix sebelum direduksi akan dibutuhkan 100000 entri. Namun, apabila dengan menggunakan PLSA (melibatkan adanya topik) dengan melakukan reduksi dimensi, maka akan diperoleh matriks dengan dimensi 100x10 dan 1000x10. Apabila jumlah entri dari kedua matriks tersebut dijumlahkan, maka akan menjadi 11000 entri. Perubahan jumlah entri dari 100000 entri menjadi 11000 entri ini yang dimaksudkan dengan reduksi dimensi pada PLSA. Reduksi dimensi tersebut membuat algoritma akan berjalan lebih efisien.

## 2.2. Classifier
## 2.2.1. Support Vector Machine (SVM)

Support Vector Machine (SVM) adalah sebuah metode klasifikasi dan regresi yang mengkombinasikan algoritma komputasional dengan hasil teoretikal; kedua karakteristik ini memberikannya reputasi yang bagus dan menaikkan pamornya dalam penggunaannya di berbagai area. (Cortes and Vapnik, 1995).

SVM merupakan sebuah teknik yang baru yang cocok untuk *binary classification task*, yang terkait dan memuat elemen-elemen statistik terapan non parameterik, jaringan saraf tiruan, dan *machine learning*. (Auria and Moro, 2008).

## 2.2.2. Logistic Regression

Penggunaan *logistic regression* untuk memprediksi class probability adalah sebuah pemilihan pemodelan, sama seperti pemilihan pemodelan untuk memprediksi variabel kuantitatif dengan menggunakan *linear regression*. *Logistic regression* adalah salah satu dari tool yang paling umum digunakan pada statistik terapan dan analisis data diskrit (Shalizi, 2012).

## 3. Metodologi

PLSA dapat diterapkan ke dalam berbagai aplikasi seperti penilai esai otomatis, peringkas dokumen, dan lain sebagainya. Pada penelitian ini, yang dikerjakan oleh algoritma PLSA adalah hanya terbatas pada pekerjaan untuk melakukan klasifikasi dokumen teks. Dari sejumlah data training yang sudah disediakan, maka algoritma EM akan menjalankan proses training. Proses training akan dilakukan dengansejumlah angka iterasi tertentu. Output dari algoritma EM merupakan model dari hasil training yang dilakukan oleh PLSA. Perlakuan yang sama akan ditujukan pada data testing. Dari data testing yang sudah disediakan, akan diambil representasi dokumen yang nantinya akan dibandingkan dengan model PLSA. Dengan menggunakan *classifier*, maka akan dihitung keakuratan metode PLSA dalam melakukan klasifikasi dokumen teks.

Data yang digunakan merupakan korpus yang merupakan koleksi dari artikel berita berbahasa Indonesia dari Kompas, Tempo, dan Republika. Korpus dibagi menjadi 4 kategori yang khusus yaitu ekonomi, internasional, politik, dan olahraga.

### 3.1. Document Preprocessing

*Document preprocessing* merupakan proses pengolahan dokumen ke dalam bentuk yang lebih padat, dimana bentuk tersebut mewakili makna dari dokumen secara utuh. Keseluruhan data yang digunakan akan melewati tahap document preprocessing terlebih dahulu.

1. Tokenisasi

   Tokenisasi merupakan proses pemotongan dari bentuk kalimat menjadi kumpulan kata-kata yang disebut sebagai token. Token akan digunakan sebagai representasi data pada tiap baris yang ada di matriks probabilitas.

2. Pembuangan stop word

   Sebelum token yang sudah dihasilkan dari proses tokenisasi digunakan pada matriks probabilitas, akan diproses terlebih dahulu token mana saja yang merupakan stop word untuk kemudian dihilangkan. *Stop word* merupakan kata yang tidak mewakili makna dari suatu konteks. Beberapa contoh dari *stop word* pada bahasa Indonesia adalah "yang", "dari", "kemudian" dan lain sebagainya. Hasil dari pembuangan stop word ini yang kemudian akan menjadi representasi baris pada matriks probabilitas, baik pada data *training* ataupun data *testing*.

### 3.2. Proses Training Data

Terdapat kurang lebih 1000 dokumen yang akan digunakan untuk melakukan *training* pada model PLSA. Data yang disediakan untuk *training* tersebut akan divariasikan menjadi 400, 700, dan 1000 dokumen. Hal ini dilakukan untuk mengamati perbedaan yang terjadi apabila jumlah dokumen yang digunakan untuk *training* berbeda. 1000 dokumen tersebut terdiri dari masing-masing 250 dokumen dari 4 kategori, yaitu ekonomi, internasional, politik, dan olahraga. Berita tersebut diambil dari kumpulan artikel Tempo, Kompas, dan Republika.

Supaya lebih bisa menarik kesimpulan dari berbagai macam parameter dan konfigurasi, maka dibuat skenario untuk pelatihan data sebagai berikut:

1. Variasi dengan menggunakan jumlah topik yang akan dihasilkan (3, 4, dan 5 topik)
2. Variasi dengan jumlah iterasi yang berbeda (1, 3, 5, 7, 9, 10 dan 20 iterasi)
3. Variasi jumlah data *training* (400, 700, dan 1000 dokumen)
4. Perulangan konfigurasi yang sama sebanyak 2 kali

*Training* dengan menggunakan algoritma EM dilakukan pada matriks $P(d|z)$ yang dilatih sesuai dengan jumlah topik, jumlah iterasi, jumlah data training dan perulangan konfigurasi yang sama sesuai korpus yang sudah tersedia. Matriks $P(d|w)$ tidak dimasukkan ke dalamnya karena yang sedang diuji adalah mengenai seberapa efektif reduksi dimensi yang dilakukan oleh PLSA (seperti dijelaskan pada bagian 2), sehingga hanya $P(z|d)$ saja yang terlibat.

Setelah dilakukan pemodelan menggunakan algoritma EM, maka diperoleh matriks yang baru. Matriks tersebut merupakan model yang sudah terbentuk dari pelatihan dari sejumlah data. Pada saat *training*, setiap dokumen yang diproses akan diberikan identitas khusus supaya ketika proses klasifikasi dokumen berlangsung, semuanya bisa berjalan dengan baik. Contoh identitas khusus adalah apabila diambil data dari kategori ekonomi, maka pada saat training data selesai dijalankan, dokumen tersebut tetap dikenali sebagai kategori ekonomi. Caranya adalah dengan memberikan penanda pada dokumen tersebut.

### 3.3. Proses Testing Data

Data yang disediakan untuk keperluan uji coba berjumlah 100 dokumen, masing-masing berjumlah 25 dokumen. Baik kategori ekonomi, internasional, politik ataupun olahraga jumlahnya adalah sama. Berita tersebut juga diambil dari kumpulan artikel Tempo, Kompas, dan Republika.

Mekanisme yang dibuat mirip dengan data *training*. Dari dokumen *testing* yang disediakan, akan diekstrak topik-topiknya. Matriks yang terbentuk akan terdiri dari banyak matriks probabilitas. Namun, karena pada data *testing* tidak menggunakan algoritma PLSA, maka ekstraksi topiknya adalah dilakukan dengan mencocokkan kata-kata yang ada di dalam dokumen *testing* dengan kata-kata yang ada pada model hasil dari *training* data. Pasti akan ditemukan kata-kata yang ada pada *training* data dan ada juga yang tidak terdapat pada *training* data. Apabila kata-kata tersebut memiliki kesamaan dengan kata yang ada pada *training* data, maka kata tersebut akan dinilai sebagai topik dari dokumen *testing*. Nilai probabilitas yang diperoleh juga datang dari proses lookup kepada matriks probabilitas dari masing-masing kata yang disebut sebagai topik dari dokumen *testing*. Sama halnya seperti data *training*, setiap dokumen yang diproses akan diberikan identitas dari mana kategori asli dari data tersebut. Informasi ini yang nanti digunakan sebagai pembanding apakah algoritma PLSA berhasil untuk menghasilkan representasi dokumen dengan nilai akurasi yang baik. Hasil dari pengolahan dokumen *testing* itu akan menghasilkan sebuah matriks P(z|d). Matriks yang dihasilkan dari hasil *lookup* kurang lebih bisa dicontohkan pada gambar 3.

$$\begin{matrix} w17 \\ w48 \\ w211 \\ w286 \end{matrix} \begin{pmatrix} z1 & ... & ... & zk \\ z1 & ... & ... & zk \\ & & & \\ & & & \end{pmatrix}$$

**Gambar 3. Matriks dari Testing Documents**

Pada gambar 3, diasumsikan bahwa kata ke-17, ke-48, ke-211, dan ke-286 merupakan kata-kata yang muncul pada *training documents*, sehingga matriks tersebut dibentuk dari nilai probabilitas masing-masing kata. Dari keseluruhan nilai probabilitas $z_1$ hingga $z_k$ pada semua kata yang muncul di *training documents*, selanjutnya akan dihitung nilai rata-ratanya dan menghasilkan satu vektor probabilitas yang mewakili dokumen testing.

### 3.4. Pengujian

Pengujian terhadap algoritma PLSA ini dikerjakan dengan melibatkan *classifier*. *Classifier* yang digunakan adalah Support Vector Machine (SVM) dan Logistic Regression. Data hasil *training* yang merupakan model PLSA akan dijadikan patokan di dalam penilaian akurasi algoritma. Dari semua *testing documents* yang sudah diproses menjadi representasi matriks akan diukur dengan menggunakan *classifier*. Seperti sebelumnya sudah disebutkan bahwa setiap matriks pada *testing documents* memiliki identitas khusus yang berisi informasi dari mana kategorinya.

Di sisi lain, pada saat *classifier* melakukan proses pada *testing documents* dengan berpatokan pada model data *training*, akan dihasilkan output yang menginformasikan dokumen tersebut masuk ke dalam kategori yang mana. Perbandingan antara kedua data ini yang nantinya diukur satu persatu sehingga akan diperoleh nilai akurasi dari algoritma PLSA dalam melakukan klasifikasi dokumen teks.

## 4. Implementasi

Implementasi dari metode PLSA diterapkan menggunakan Python *programming language*. Sedangkan untuk melakukan klasifikasi dokumen, software yang digunakan adalah Weka 3.6. Di dalam Weka ini terdapat banyak *classifier* yang bisa digunakan untuk pengolahan data. Dari banyaknya opsi yang diberikan oleh Weka, maka Support Vector Machine (libSVM) dan Logistic Regression (Logistic) yang digunakan sebagai *classifier* di penelitian ini.

Sesuai dengan keterangan sebelumnya, korpus yang digunakan merupakan koleksi dari artikel berita berbahasa Indonesia yang diperoleh dari Kompas, Tempo, dan Republika. Korpus dibagi menjadi 4 kategori yaitu ekonomi, internasional, politik, dan olahraga. Berbagai variasi konfigurasi untuk testing dilakukan supaya bisa ditarik kesimpulan dari berbagai perspektif. Bagaimana perbandingan hasil akurasi dari classifier SVM dan Logistic, kemudian bagaimana hasilnya jika *training* dilakukan hanya dengan 5 iterasi dan 20 iterasi, dan lain sebagainya.

## 5. Hasil dan Pembahasan

Setelah dilakukan eksperimen secara lebih komprehensif dari keseluruhan variasi yang ditetapkan. Eksperimen dengan konfigurasi yang sama dilakukan sebanyak 2 kali. Hasil akurasi yang diperoleh dari kedua macam uji coba tersebut bisa dilihat pada tabel 1 dan 2.

Tabel 1. Hasil Uji Coba pada Percobaan Pertama

|  | Eksperimen I | | | | | | | | | | | | | | | | | |
|---|---|---|---|---|---|---|---|---|---|---|---|---|---|---|---|---|---|---|
|  | 400 dokumen | | | | | | 700 | | | | | | 1000 | | | | | |
|  | 3 topik | | 4 topik | | 5 topik | | 3 topik | | 4 topik | | 5 topik | | 3 topik | | 4 topik | | 5 topik | |
|  | SVM | Log | SVM | Log | SVM | Log | SVM | Log | SVM | Log | SVM | Log | SVM | Log | SVM | Log | SVM | Log |
| **1 iterasi** | 36% | 40% | 47% | 41% | 35% | 36% | 26% | 43% | 20% | 29% | 33% | 33% | 40% | 34% | 38% | 39% | 42% | 34% |
| **3 iterasi** | 43% | 42% | 41% | 59% | 35% | 48% | 45% | 37% | 45% | 48% | 38% | 55% | 37% | 40% | 29% | 35% | 28% | 42% |
| **5 iterasi** | 45% | 57% | 39% | 53% | 58% | 66% | 29% | 41% | 49% | 53% | 59% | 60% | 31% | 47% | 51% | 43% | 40% | 48% |
| **7 iterasi** | 36% | 51% | 49% | 68% | 45% | 57% | 33% | 69% | 55% | 66% | 68% | 77% | 39% | 56% | 43% | 62% | 33% | 61% |
| **9 iterasi** | 31% | 43% | 37% | 63% | 39% | 63% | 37% | 60% | 45% | 68% | 60% | 76% | 54% | 64% | 35% | 57% | 42% | 62% |
| **10 iterasi** | 44% | 61% | 69% | 64% | 41% | 63% | 57% | 74% | 58% | 68% | 42% | 62% | 44% | 59% | 61% | 68% | 55% | 71% |
| **20 iterasi** | 52% | 60% | 63% | 62% | 59% | 68% | 55% | 62% | 69% | 71% | 66% | 76% | 61% | 68% | 65% | 69% | 67% | 68% |

**Tabel 2. Hasil Uji Coba pada Percobaan Kedua**

| | Eksperimen II ||||||||||||||||||
|---|---|---|---|---|---|---|---|---|---|---|---|---|---|---|---|---|---|---|
| | 400 dokumen |||||| 700 |||||| 1000 ||||||
| | 3 topik || 4 topik || 5 topik || 3 topik || 4 topik || 5 topik || 3 topik || 4 topik || 5 topik ||
| | SVM | Log | SVM | Log | SVM | Log | SVM | Log | SVM | Log | SVM | Log | SVM | Log | SVM | Log | SVM | Log |
| **1 iterasi** | 43% | 37% | 27% | 24% | 24% | 35% | 31% | 39% | 25% | 33% | 42% | 45% | 30% | 32% | 31% | 33% | 35% | 44% |
| **3 iterasi** | 40% | 50% | 37% | 38% | 34% | 41% | 36% | 46% | 50% | 55% | 48% | 61% | 39% | 48% | 42% | 47% | 37% | 41% |
| **5 iterasi** | 46% | 53% | 46% | 63% | 39% | 66% | 26% | 39% | 42% | 53% | 38% | 52% | 33% | 45% | 42% | 56% | 30% | 43% |
| **7 iterasi** | 66% | 72% | 50% | 56% | 49% | 57% | 26% | 44% | 29% | 54% | 71% | 71% | 25% | 47% | 43% | 60% | 26% | 49% |
| **9 iterasi** | 60% | 67% | 55% | 51% | 51% | 77% | 34% | 57% | 31% | 59% | 63% | 75% | 41% | 57% | 50% | 61% | 33% | 48% |
| **10 iterasi** | 33% | 54% | 67% | 67% | 35% | 51% | 58% | 63% | 59% | 68% | 73% | 70% | 38% | 57% | 43% | 60% | 50% | 61% |
| **20 iterasi** | 40% | 59% | 75% | 81% | 55% | 72% | 54% | 75% | 65% | 74% | 55% | 68% | 44% | 43% | 69% | 84% | 77% | 82% |

Berdasarkan data yang disediakan pada tabel 1 dan 2, dapat diukur beberapa hal sebagai perbandingan yang komprehensif mengenai variasi jumlah topik, variasi *classifier*, variasi jumlah dokumen training, dan variasi jumlah iterasi.

## 5.1. Jumlah Topik

Dari masing-masing jumlah topik pada berbagai konfigurasi yang sudah dibuat maka akan dihitung rata-rata akurasinya. Data hasil perhitungan rata-rata dari masing-masing jumlah topik adalah:

- Jumlah topik 3 : 47%
- Jumlah topik 4 : 52%
- Jumlah topik 5 : 52%

Dengan variasi jumlah topik diperoleh bahwa akurasi tertinggi ada pada jumlah topik 4 dan 5.

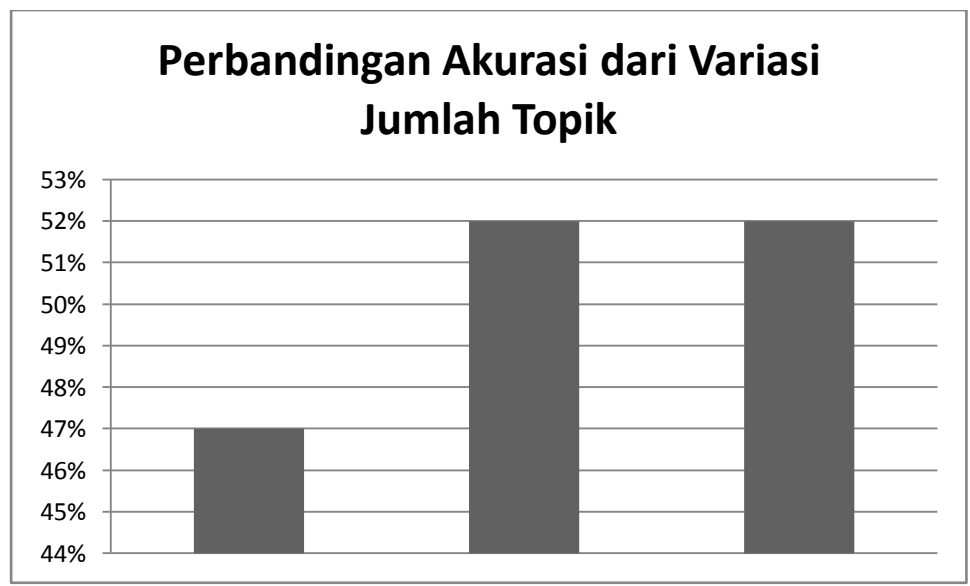

**Gambar 4. Perbandingan Akurasi dari Variasi Jumlah Topik**

### 5.2. Jenis Classifier

Dari dua jenis *classifier* yang digunakan pada berbagai konfigurasi yang sudah dibuat maka akan dihitung rata-rata akurasinya. Data hasil perhitungan rata-rata dari masing-masing *classifier* adalah:

- SVM Classifier: 45%
- Logistic Regression Classifier: 55%

Bisa dilihat data secara statistik bahwa classifier Logistic Regression memberikan kinerja yang lebih baik dari *classifier* Support Vector Machine (SVM).

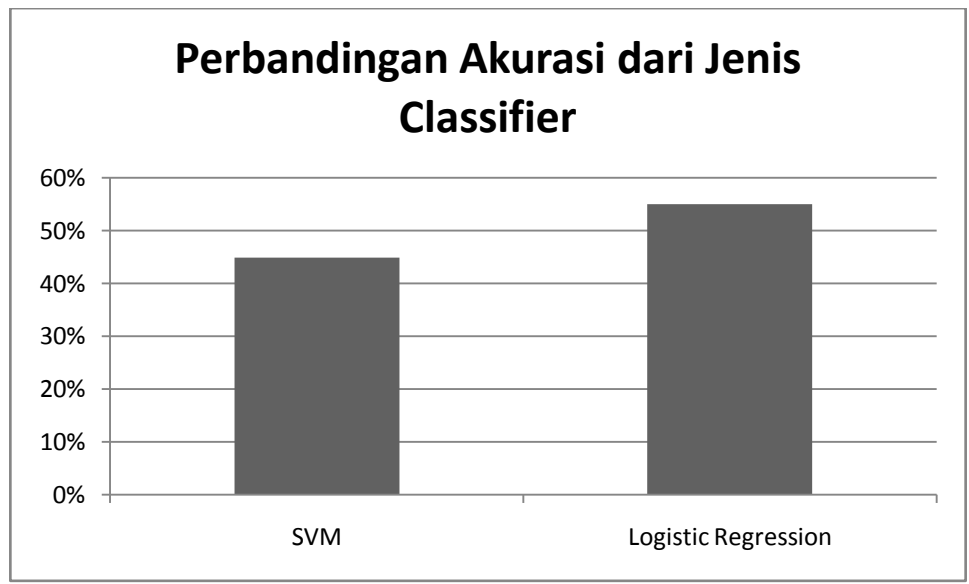

**Gambar 5. Perbandingan Akurasi dari Jenis Classifier**

### 5.3. Jumlah Dokumen Training

Dari variasi jumlah dokumen *training* pada berbagai konfigurasi yang sudah dibuat maka akan dihitung rata-rata akurasinya. Data hasil perhitungan rata-rata dari berbagai variasi jumlah dokumen training adalah:

- 400 dokumen *training*: 51%
- 700 dokumen *training*: 52%
- 1000 dokumen *training*: 48%

Bisa dilihat bahwa penambahan jumlah dokumen untuk data *training* mengindikasikan tidak adanya pengaruh secara signifikan pada hasil akurasi. Dari eksperimen yang dilakukan, hasil yang ditunjukkan pada training dengan menggunakan data yang lebih banyak cenderung sama dan justru lebih kecil dari data yang lebih sedikit.

Melihat kecenderungan yang tidak umum pada hasil data ini yakni dengan bertambahnya training data seharusnya akurasi akan lebih meningkat. Hipotesis yang muncul karena hal ini adalah bahwa jumlah data pada setiap sumber artikel tidak seimbang, sehingga model yang terbentuk tidak proporsional atau seimbang untuk mewakili berbagai gaya bahasa yang ada pada ketiga media tersebut yaitu Kompas, Tempo, dan Republika.

Akan tetapi hal tersebut masih harus diuji kembali supaya terbukti validitasnya.

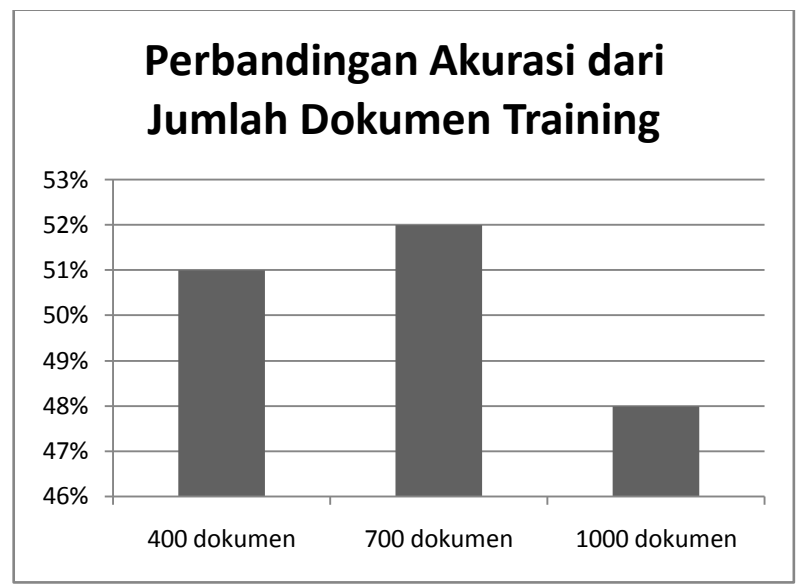

**Gambar 6. Perbandingan Akurasi dari Jumlah Dokumen Training**

### 5.4. Jumlah Iterasi

Dari variasi jumlah iterasi pada berbagai konfigurasi yang sudah dibuat maka akan dihitung rata-rata akurasinya. Data hasil perhitungan rata-rata dari berbagai variasi jumlah iterasi adalah seperti terlihat pada tabel 3.

**Tabel 3. Data dan Rata-rata dari Variasi Jumlah Iterasi**

| Jumlah Iterasi | Rata-rata |
|---|---|
| 1 iterasi | 35% |
| 3 iterasi | 43% |
| 5 iterasi | 47% |
| 7 iterasi | 52% |
| 9 iterasi | 53% |
| 10 iterasi | 57% |
| 20 iterasi | 65% |

Dari pengamatan pada hasil iterasi yang berbeda dari data yang sama, diperoleh hasil bahwa nilai akurasi mengindikasikan ciri-ciri meningkat.

## 6. Kesimpulan

Dari beberapa variasi yang dilakukan pada data hasil *training* dan *testing*, dapat disimpulkan beberapa hal sebagai berikut:

1. Jumlah topik yang didefinisikan sebagai output dari algoritma PLSA sangat dipengaruhi oleh topik yang didefinisikan. Apabila beberapa topik yang didefinisikan merupakan topik yang cukup beririsan maka nilai akurasi dari variasi jumlah topik tidak akan jauh berbeda.
2. Classifier Logistic memiliki performa yang lebih baik dari classifier SVM. Hal ini sesuai dengan data yang digunakan untuk uji coba. Karena data yang digunakan merupakan angka diskrit, klasifikasi dengan Logistic Regression memberikan hasil yang lebih baik daripada SVM. Hal ini sesuai dengan teorinya bahwa SVM lebih bagus untuk klasifikasi data untuk bilangan real.
3. Jumlah dokumen *training* seharusnya berpengaruh signifikan untuk uji coba penggunaan algoritma PLSA, akan tetapi cara bagaimana data *training* dan *testing* disiapkan butuh menjadi perhatian yang khusus.
4. Semakin banyak iterasi yang dilakukan maka nilai akurasi akan terus meningkat. Hal ini akan terus berlanjut sampai ditemukan kondisi konvergen dimana nilai akurasi tidak akan meningkat lagi

## 7. Referensi